\pdfoutput=1

\documentclass[11pt]{article}

\usepackage[final]{acl}
\usepackage{times}
\usepackage{latexsym}

\usepackage[T1]{fontenc}

\usepackage[utf8]{inputenc}

\usepackage{microtype}

\usepackage{amsmath}
\usepackage{booktabs}
\usepackage{hyperref}
\usepackage{enumitem}
\setlist{leftmargin=*,nosep}
\definecolor{headergray}{gray}{0.8}
\definecolor{sectionblue}{RGB}{173,216,230}
\usepackage{tcolorbox}
\usepackage{cleveref}

\usepackage{inconsolata}

\usepackage{graphicx}

\title{LegalSeg: Unlocking the Structure of Indian Legal Judgments Through Rhetorical Role Classification}

\author{Shubham Kumar Nigam$^{1}$ \qquad Tanmay Dubey$^{1}$ \qquad  Govind Sharma$^{1}$ \\ 
\textbf{Noel Shallum}$^{3}$ \qquad \textbf{Kripabandhu Ghosh}$^{2}$ \qquad \textbf{Arnab Bhattacharya}$^{1}$\\
$^{1}$ IIT Kanpur, India \quad
$^{2}$  IISER Kolkata, India \quad
$^{3}$ Symbiosis Law School Pune, India \\
\texttt{\{sknigam, tanmay, govind, arnabb\}@cse.iitk.ac.in} \\
\texttt{kripaghosh@iiserkol.ac.in,} \quad \texttt{noelshallum@gmail.com}
}

\begin{document}
\maketitle

\begin{abstract}
In this paper, we address the task of semantic segmentation of legal documents through rhetorical role classification, with a focus on Indian legal judgments. We introduce \texttt{LegalSeg}, the largest annotated dataset for this task, comprising over 7,000 documents and 1.4 million sentences, labeled with 7 rhetorical roles. To benchmark performance, we evaluate multiple state-of-the-art models, including Hierarchical BiLSTM-CRF, TransformerOverInLegalBERT (ToInLegalBERT), Graph Neural Networks (GNNs), and Role-Aware Transformers, alongside an exploratory \texttt{RhetoricLLaMA}, an instruction-tuned large language model. Our results demonstrate that models incorporating broader context, structural relationships, and sequential sentence information outperform those relying solely on sentence-level features. Additionally, we conducted experiments using surrounding context and predicted or actual labels of neighboring sentences to assess their impact on classification accuracy. Despite these advancements, challenges persist in distinguishing between closely related roles and addressing class imbalance. Our work underscores the potential of advanced techniques for improving legal document understanding and sets a strong foundation for future research in legal NLP.
\end{abstract}



\section{Introduction}
The increasing complexity of legal documents necessitates the use of advanced NLP techniques to aid in their understanding and analysis. Semantic segmentation of legal texts into rhetorical roles is essential for improving the efficiency of legal research, enhancing access to justice, and supporting automated legal decision-making systems. It also facilitates various downstream tasks, such as legal search, summarization, and case analysis. Traditional methods often struggle with the intricacies of legal language, making it imperative to develop models that can accurately classify and interpret these documents. This paper addresses the challenge of semantic segmentation in legal documents, with a focus on the Indian judiciary’s legal judgments. Historically, the lack of large-scale annotated datasets has hindered the effective training of state-of-the-art ML models in this domain.

Previous research in this domain has highlighted the importance of annotated datasets for training effective models. However, many existing studies have relied on relatively small annotated datasets, limiting their applicability and effectiveness in real-world scenarios. For instance, datasets such as those compiled by \citet{bhattacharya2019identification, kalamkar-etal-2022-corpus} and \citet{malik-etal-2022-semantic} provided valuable insights but were constrained in size, thereby restricting the scope of their findings. In contrast, this study leverages a newly compiled dataset, \texttt{LegalSeg}, which consists of 7,120 annotated legal documents and 14,87,149 sentences. This dataset is considerably larger than those used in previous research, particularly in terms of its volume and diversity, as illustrated in Table \ref{table:legal_corpora_overview}, which summarizes various legal corpora for rhetorical role classification.
\begin{table}[t]
\centering
\resizebox{\columnwidth}{!}{%
\begin{tabular}{|l|l|l|c|c|c|l|}
\hline
\textbf{Corpus} & \textbf{Country} & \textbf{Language} & \textbf{\# Cases} & \begin{tabular}[c]{@{}c@{}}\textbf{Total \#} \\ \textbf{Sentences}\end{tabular} & \begin{tabular}[c]{@{}c@{}}\textbf{\# Labels} \end{tabular} & \textbf{Domain Coverage} \\ \hline

\citet{bhattacharya2019identification} & India & English & 50 & 9,380 & 7 & Supreme Court \\ \hline

\citet{majumder2020rhetorical} & India & English & 60 & - & 7 & \begin{tabular}[l]{@{}l@{}}Supreme Court\\High Courts\\Tribunal Courts\end{tabular} \\ \hline

\citet{malik-etal-2022-semantic} & India & English & 100 & 21,184 & 13 & \begin{tabular}[l]{@{}l@{}}Supreme Court\\Bombay High Court\\Kolkata High Court\end{tabular} \\ \hline

\citet{kalamkar-etal-2022-corpus} & India & English & 354 & 40,305 & 13 & \begin{tabular}[l]{@{}l@{}}Supreme Court\\High Courts\\District Courts\end{tabular}  \\ \hline

\citet{marino2023automatic} & India & English & 275 & 31,865 & 13 & \begin{tabular}[l]{@{}l@{}}Supreme Court\\High Courts\\District Courts\end{tabular} \\ \hline

\citet{marino2023automatic} & Italy & Italian & 1,488 & 95,920 & 5 & \begin{tabular}[l]{@{}l@{}}Civil Law of\\Italian Courts \end{tabular}  \\ \hline

\citet{modi-etal-2023-semeval} & India & English & 265 & 26,304 & 13 & Not Mentioned \\ \hline

\textbf{\texttt{LegalSeg} (Ours)} & India & English & 7,120 & 14,87,149 & 7 & \begin{tabular}[l]{@{}l@{}}Supreme Court\\High Courts\end{tabular} \\ \hline
\end{tabular}%
}
\caption{Overview of Legal Corpora for Rhetorical Role Classification}
\label{table:legal_corpora_overview}
\end{table}
\begin{figure*}[t]
    \centering
    \includegraphics[width=\linewidth]{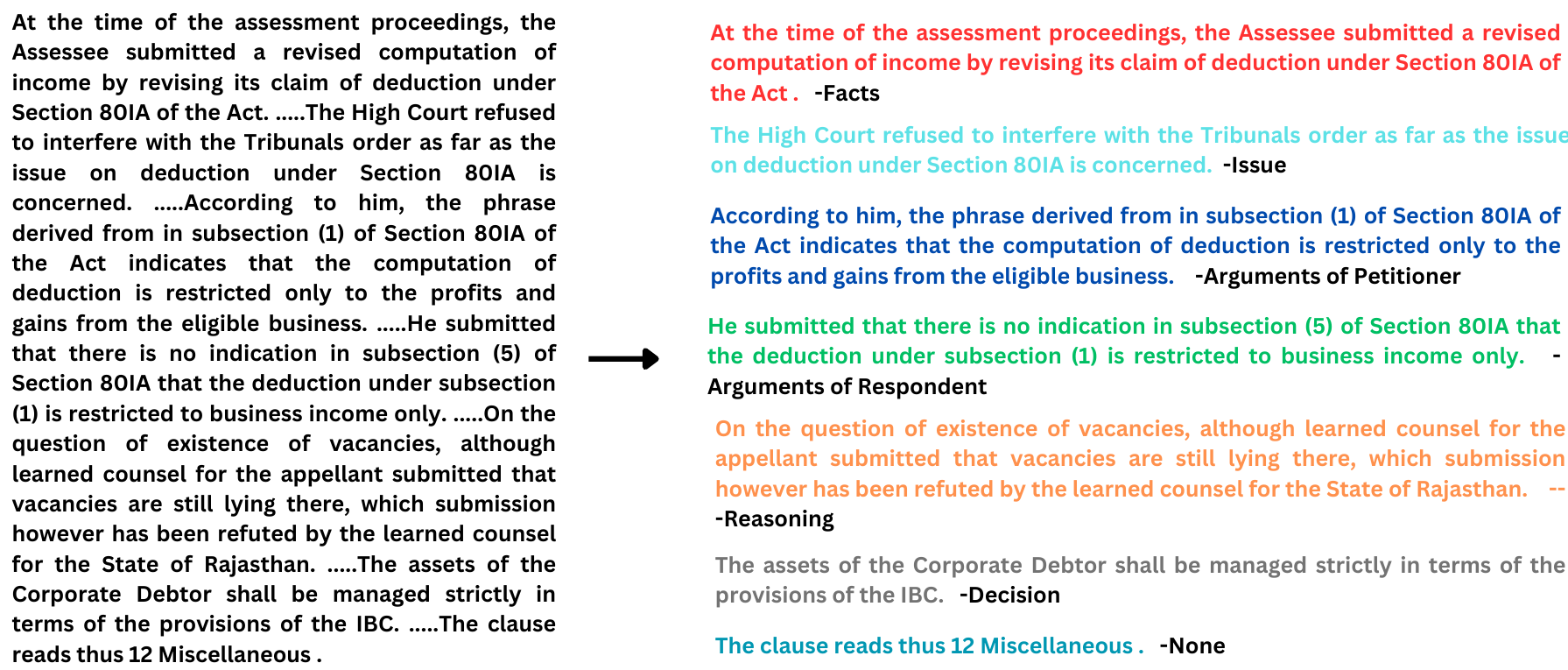}
    \caption{Example illustrating document segmentation using rhetorical roles. The left side shows an excerpt from a legal document, while the right side demonstrates the segmentation and labeling of sentences.}
    \label{fig:RR_examples}
\end{figure*}
An example of how a legal judgment is semantically segmented into rhetorical roles is illustrated in Figure \ref{fig:RR_examples}. As shown, an unstructured legal document is broken down into coherent parts, each annotated with a rhetorical role label such as Facts, Reasoning, or Decision. This segmentation is critical for understanding the flow of arguments and supporting the automation of legal processes.

We implemented several SoTA models to evaluate the effectiveness of our dataset. Among these, the Hierarchical BiLSTM-CRF model \citet{bhattacharya2019identification} captures contextual information using a hierarchical approach, while MultiTask Learning (MTL) incorporates label shift prediction to refine the identification of rhetorical roles by considering role transitions \citet{malik-etal-2022-semantic}. Additionally, we explored LEGAL-TransformerOverBERT (LEGALToBERT), a hierarchical architecture stacking a transformer encoder over a legal-domain-specific BERT model, which effectively captures sentence relationships and positional encoding within legal documents \citet{marino2023automatic}.

In addition to these models, we introduce novel approaches, including InLegalToBERT, Graph Neural Networks (GNNs), and Role-Aware Transformers, mostly that have not been previously explored in the context of rhetorical role classification in legal texts. InLegalToBERT, a variant of LEGALToBERT, incorporates the total number of sentences as an additional feature to enhance the model’s ability to capture positional information within documents. GNNs leverage the structural relationships between sentences by representing them as nodes in a graph, allowing for effective propagation of information across sentence pairs and capturing both local and global context. Role-Aware Transformers, on the other hand, utilize specialized embeddings to incorporate rhetorical role-specific information into pre-trained models, improving the model’s ability to differentiate between closely related roles.

A key focus of our work is on the use of open-source large language models (LLMs), which align with the principles of accessibility and reproducibility in research. Instead of leveraging proprietary models like GPT-4, which are costly and lack transparency, we explore the potential of open-source models fine-tuned for legal NLP tasks. Specifically, we developed and investigated \texttt{RhetoricLLaMA}, a fine-tuned version of the open-source LLaMA-2-7B architecture, designed for semantic segmentation in legal documents. While the initial performance of \texttt{RhetoricLLaMA} was lower than anticipated, it highlights both the promise and the challenges of instruction-tuned LLMs for handling complex legal language. Given the computational limitations, our approach ensures that our models remain accessible for broader research communities, facilitating reproducibility without incurring significant costs. 

Our contributions to this work are as follows:
\begin{enumerate}
    \item Introduction of the \texttt{LegalSeg} dataset, the largest annotated dataset for rhetorical role classification in legal documents.
    \item Implementation and evaluation of SoTA models for semantic segmentation of legal texts.
    \item The development of novel models, including InLegalToBERT, Graph Neural Networks (GNNs), and Role-Aware Transformers, which enhance representation and context handling for rhetorical role classification.
    \item Exploration of instruction-tuned LLMs, through the development of \texttt{RhetoricLLaMA}, highlighting the potential and limitations of LLMs in rhetorical role classification.
\end{enumerate}

To ensure reproducibility, we have made the \texttt{LegalSeg} dataset and the code for all our models accessible via a GitHub link\footnote{\href{https://github.com/ShubhamKumarNigam/LegalSeg}{https://github.com/ShubhamKumarNigam/LegalSeg}}.

\section{Related Work}
Recent advancements in legal text processing have spurred significant research efforts aimed at automating various tasks such as semantic segmentation, judgment prediction, and summarization of legal documents. However, much of this work relies heavily on manual annotation, with many studies focusing on the intricacies of annotation processes, including the development of annotation guidelines, IAA studies, and the curation of gold standard corpora. For instance, the TEMIS corpus, which consists of 504 sentences annotated both syntactically and semantically, was developed to enhance understanding of legislative texts \citet{venturi2012design}. Additionally, an in-depth annotation study highlighted low assessor agreement for labels such as Facts and Reasoning \citet{wyner2013case}. In the Indian context, datasets like ILDC \citet{malik-etal-2021-ildc}, PredEx \citet{nigam-etal-2024-legal} and \citet{nigam2022nigam, malik-etal-2022-semantic, nigam2023nonet, nigam2023legal} have highlighted the growing role of AI in legal judgments, with an emphasis on explainability. Research in LJP with LLMs, such as \citet{vats2023llms} and \citet{nigam-etal-2024-legal}, has experimented with models like GPT-3.5 Turbo and LLaMA-2 on Indian legal datasets. 

Several efforts have been made to automate the annotation task itself. For example, \citet{wyner2010towards} discusses methodologies that employ NLP tools to analyze 47 criminal cases from California courts. Initial experiments aimed at understanding rhetorical roles within court documents were often intertwined with broader goals of document summarization \citet{saravanan2008automatic}. 

Further contributions include segmenting documents into functional parts (e.g., Introduction, Background) and issue-specific sections \citet{vsavelka2018segmenting}. A semi-supervised training method for identifying factual versus non-factual sentences was explored by \citet{nejadgholi2017semi} using a fastText classifier. The comparison between rule-based scripts and machine learning approaches for rhetorical role identification was conducted by \citet{walker2019automatic} demonstrating the efficacy of both methodologies in this context.

In recent studies, \citet{bhattacharya2019identification} proposed a CRF-BiLSTM model specifically for assigning rhetorical roles to sentences in Indian legal documents \citet{bhattacharya2019identification, malik-etal-2022-semantic} created a comprehensive rhetorical role corpus annotated with 13 fine-grained roles and developed a multi-task learning model for prediction tasks. \citet{kalamkar-etal-2022-corpus} constructed a corpus consisting of 354 Indian legal documents annotated with rhetorical roles across 40,305 sentences and introduced a transformer-based baseline model.

Moreover, \citet{malik-etal-2022-semantic} proposed an MTL framework that significantly improved classification scores by leveraging a Hierarchical BiLSTM with CRF architecture. \citet{marino2023automatic} introduced LEGAL-ToBERT, which integrates a transformer encoder atop a legal-domain-specific BERT model tailored for both Italian and Indian datasets. More recently, the HiCuLR framework \citet{santosh2024hiculr} introduced hierarchical curriculum learning for rhetorical role labeling, progressively training models with a structured, easy-to-difficult learning strategy, which enhances performance across multiple rhetorical role datasets.

\section{Task Description}
The goal of this research is to develop models capable of performing semantic segmentation on legal documents by identifying and classifying rhetorical roles (RR) within the text. Let \( D = \{d_1, d_2, \dots, d_n\} \) represent a collection of legal documents, where \( d_i \in D \) consists of a sequence of sentences \( S_i = \{s_{i1}, s_{i2}, \dots, s_{im}\} \), with \( m \) representing the number of sentences in document \( d_i \). The task is to assign a rhetorical role label \( y_{ij} \in Y \) to each sentence \( s_{ij} \), where $Y$ is the predefined set of 7 rhetorical role labels.

Formally, the task can be described as:
\[
f: S_i \rightarrow Y
\]
\[
Y = \left\{
\begin{aligned}
&\text{Facts}, \, \text{Issue}, \, \text{Arguments of Petitioner}, \\
&\text{Arguments of Respondent}, \, \text{Reasoning},  \\
&\text{Decision}, \, \text{None}
\end{aligned}
\right\}
\]
where \( f \) is a function that maps each sentence \( s_{ij} \) in a document \( d_i \) to its corresponding rhetorical role label \( y_{ij} \). Thus, the goal is to find:
\[
f(s_{ij}) = y_{ij}, \quad \forall s_{ij} \in S_i, \quad y_{ij} \in Y
\]
The input to the system is a legal document \( d_i \), and the output is a sequence of rhetorical role labels corresponding to each sentence in the document:
\[
f(S_i) = \{y_{i1}, y_{i2}, \dots, y_{im}\}, \quad y_{ij} \in Y
\]


\section{Dataset}
\begin{figure}[t] 
    \centering 
    \includegraphics[width=\linewidth]{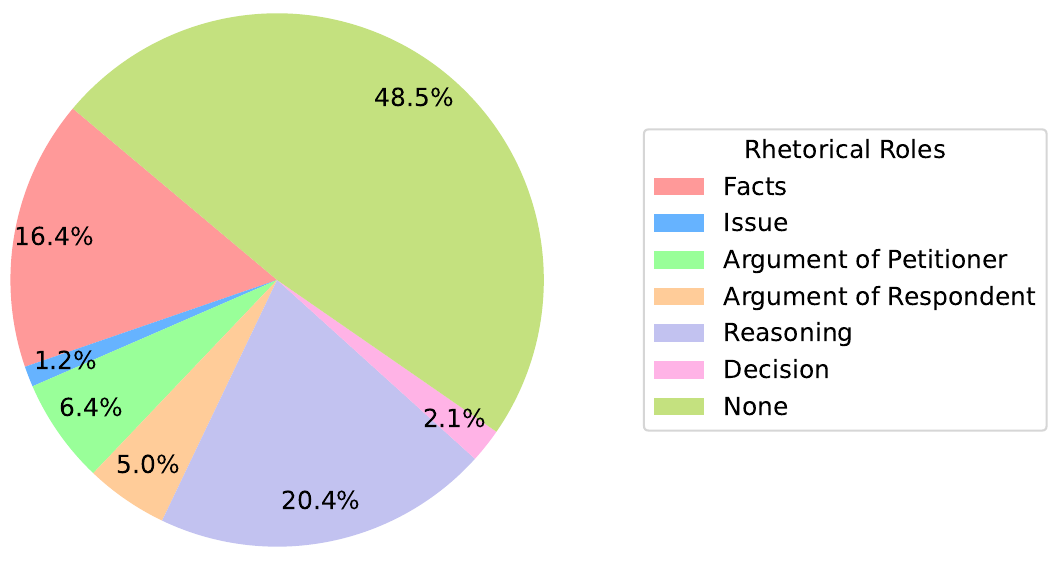} 
    \caption{Distribution of Rhetorical Roles within the Dataset.}
    \label{pie-chart}
\end{figure}
In this research, we present the \texttt{LegalSeg} Judgment Dataset, the largest annotated dataset of legal judgments in the English language, specifically focused on rhetorical role segmentation. This dataset represents a significant advancement in the field of legal Natural Language Processing (L-NLP), especially in the context of the Indian judiciary. It aims to address existing gaps in annotation comprehensiveness by offering a rich resource of annotated legal judgments designed to facilitate semantic labeling task.

\subsection{Dataset Compilation}
The dataset comprises 16,000 legal judgments sourced from the IndianKanoon database, a widely used legal search engine for Indian legal documents. These judgments were collected from the Supreme Court of India and various High Courts, ensuring a diverse selection of cases across multiple domains of law, such as criminal, civil, and constitutional matters.  

During the data curation process, several documents were excluded for reasons such as corruption (e.g., containing unrecognized characters or missing segments) or being extremely short, often comprising procedural orders rather than substantive judgments. Additionally, after annotation, the final dataset was refined to 7,120 judgments by removing documents with incomplete or ambiguous annotations, ensuring a high-quality corpus that is also the largest of its kind by a significant margin.

\subsection{Dataset Preparation and Preprocessing}
To train and evaluate models for this task, the dataset was divided into training, validation, and test sets using a 70-20-10 split, which comprises 4,984, 1,424, and 712 documents correspondingly. This split ensures a robust set of data for both training and evaluating models. Additionally, we computed various statistics regarding the documents and sentences within the dataset, including the average number of sentences per document and token counts presented in Table~\ref{tab:merged_dataset_stats}. Furthermore, the distribution of rhetorical roles within the dataset is visualized in a pie chart, Figure~\ref{pie-chart}.

To improve the performance of our models, we modified the dataset by breaking the documents into individual sentences and assigning each sentence its respective label. For sentence segmentation, we utilized SpaCy\footnote{\href{https://spacy.io/api/sentencizer}{https://spacy.io/api/sentencizer}}.

\begin{table}[t]
\centering
\resizebox{\linewidth}{!}{%
\begin{tabular}{lccc}
\toprule
\textbf{Statistic}                        & \textbf{Train Set} & \textbf{Validation Set} & \textbf{Test Set} \\ 
\midrule
{\# Documents}              & 4,984             & 1,424                   & 712             \\ 
{Total \# Sentences}        & 11,22,507         & 2,93,370                 & 1,49,881           \\ 
{Avg. \# Sentences per Doc} & 225             & 206                  & 210            \\ 

{Avg. \# Tokens per Sentence} & 34             & 30                   & 32             \\ 
\midrule
\multicolumn{4}{c}{\textbf{Sentence Count per Label}} \\ 
\midrule
Facts                                     & 1,69,653            & 51,924                  & 24,909            \\ 
Issue                                     & 12,791             & 4,259                   & 1,843             \\ 
AoP             & 64,987             & 24,707                  & 14,520            \\ 
AoR             & 50,097             & 16,021                  & 9,579             \\ 
Reasoning                                 & 2,02,346            & 67,113                  & 36,689            \\ 
Decision                                  & 19,574             & 7,634                   & 3,841             \\ 
None                                      & 6,03,059            & 1,21,712                  & 58,500            \\ 
\midrule

\multicolumn{4}{c}{\textbf{Average Number of Tokens per Label}} \\ 
\midrule
Facts                                     & 34              & 33                   & 32             \\ 
Issue                                     & 41              & 42                   & 46             \\ 
AoP             & 37              & 31                   & 33             \\ 
AoR             & 38              & 35                   & 35             \\ 
Reasoning                                 & 34              & 34                   & 33             \\ 
Decision                                  & 26              & 25                   & 25             \\ 
\bottomrule
\end{tabular}
}
\caption{Dataset Statistics for \texttt{LegalSeg} Dataset}
\label{tab:merged_dataset_stats}
\end{table}
\subsection{Annotation Process}
The annotation process was performed by a group of 10 legal experts, consisting of third and fourth year law students selected for their strong academic backgrounds and familiarity with legal processes. The annotation process spanned from April 2022 to October 2023. Each annotator was assigned 10 judgments per week, ensuring detailed attention to every document.

\subsection{Quality Control}
To ensure the annotation accuracy and consistency, we implemented multiple levels of quality control:
\begin{itemize}
    \item Senior Expert Review: All disagreements in annotations were escalated to them for resolution.
    
    \item Regular Training: Annotators participated in bi-weekly training sessions, ensuring consistency in understanding and interpreting legal content. Ambiguous or difficult segments were regularly discussed and standardized.
\end{itemize}

\subsection{Annotation Roles}
Legal experts annotated each sentence with one of the following rhetorical roles:

\begin{itemize}
    \item \textbf{Facts:} Sentences that describe the sequence of events that led to the case. These typically involve details of the circumstances and actions related to the case, providing a factual narrative of the case’s background, and details about the parties involved, including key dates, events, and parties involved.

    \item \textbf{Issue:} Sentences that outline the legal issues or questions being addressed in the case. These often identify the core legal disputes or points of law that the court must resolve to make a ruling.
 
    \item \textbf{Arguments of Petitioner (AoP):} Sentences representing the arguments made by the petitioner (the party bringing the case to court). These include claims, reasoning, and justifications presented by the petitioner to support their position and persuade the court to rule in their favor.   
    
    \item \textbf{Arguments of Respondent (AoR):} Sentences that summarize the arguments made by the respondent (the party defending against the case). Like the petitioner's arguments, these statements offer counterpoints, legal interpretations, and rebuttals designed to challenge the petitioner’s claims and persuade the court to rule in the respondent’s favor.
    
    \item \textbf{Reasoning:} Sentences that provide the rationale or reasoning behind the court's decision. This includes the application of legal principles and precedents, as well as the logic that connects the facts and arguments to the final ruling. This label captures how the court justifies its decision in light of the legal issues presented.

    \item \textbf{Decision:} Sentences that reflect the final ruling or judgment of the court. This label marks the conclusion of the case, where the court issues its verdict or order, stating the outcome of the case based on its reasoning, such as granting relief, compensation, or dismissing the case.
    
    \item \textbf{None:} Sentences that do not fall under any of the defined rhetorical roles. These sentences may include procedural information, non-substantive remarks, legal jargon, or content that is not directly relevant to the legal analysis or judgment.
\end{itemize}

This annotation schema follows closely with prior works in rhetorical role segmentation, as demonstrated by the datasets used in similar research efforts such as those by \citet{bhattacharya2019identification, kalamkar-etal-2022-corpus, malik-etal-2022-semantic}.

\section{Methodology}
This section outlines the methodology employed for the task of semantic segmentation of legal documents via rhetorical roles. We implemented several SoTA methods while also exploring new techniques. These methodologies collectively aim to enhance the model's ability to understand and classify rhetorical roles in legal texts by incorporating structural, contextual, and sequential information. Each technique addresses different aspects of the complex relationships between sentences in legal documents, contributing to more accurate and context-aware classification outcomes.

\subsection{TransformerOverInLegalBERT (ToInLegalBERT)}
This pipeline is inspired by \citet{marino2023automatic}. While they employed a general-purpose BERT model, we utilized InLegalBERT, a transformer pre-trained specifically on the Indian legal domain. This substitution enhances the model’s ability to capture domain-specific nuances, resulting in improved performance. The TransformerOverInLegalBERT (ToInLegalBERT) model follows a hierarchical architecture, consisting of four main components: (i) an InLegalBERT token-level encoder, (ii) a sentence-level positional encoder, (iii) a sentence-level encoder, and (iv) a prediction layer.

The process begins by splitting the document into sentences and tokenizing them.  Each sentence is then input into the ToInLegalBERT token-level encoder, where the pooled output—specifically, the hidden representation of the [CLS] token—is extracted. These pooled outputs are subsequently fed into the positional layer to create a position-dependent encoding for each sentence within the document. The encoded representations are then passed to the sentence-level encoder, which captures the relationships between sentences in the document, and finally, these outputs are directed to the prediction layer for rhetorical role classification. This method incorporates both the local context of sentences and their position in the document, enabling better rhetorical role classification. By incorporating both the local context of sentences and their position in the document, this method enables improved rhetorical role classification by effectively modeling the hierarchical structure of legal texts.

\subsection{Hierarchical BiLSTM CRF Classifier}
We also implemented the BiLSTM-CRF model proposed by \citet{bhattacharya2019identification}, which combines Bidirectional Long Short-Term Memory (BiLSTM) with a Conditional Random Field (CRF) layer. The input to this model is sentence embeddings generated using a sent2vec model trained on Indian Supreme Court judgments. These sentence embeddings are passed through a BiLSTM model, which captures sequential dependencies between sentences. The CRF layer on top of the BiLSTM ensures that the predicted rhetorical role labels adhere to the structured nature of legal documents. This model predicts the rhetorical role for each sentence by considering the context provided by neighboring sentences.

\subsection{Multi-Task Learning (MTL)}
Inspired by the Multi-Task Learning framework proposed by \citet{malik-etal-2022-semantic}, we adopt an MTL approach where rhetorical role prediction is the main task, and label shift prediction serves as the auxiliary task. The model consists of two components: a label shift detection component and a rhetorical role prediction component. The intuition is that the label shift between sentences (indicating a change in rhetorical role) helps improve role classification. The label shift detection component predicts whether a shift in rhetorical role occurs at the \(i^{th}\) sentence, while the rhetorical role classification component predicts the rhetorical role for that sentence. The output from both components is concatenated and passed to the CRF layer for final role predictions. The overall loss function for the MTL model is:
$L = \lambda L_{\text{shift}} + (1-\lambda) L_{\text{RR}},$
where \(L_{\text{shift}}\) corresponds to label shift prediction, \(L_{\text{RR}}\) corresponds to rhetorical role classification, and \(\lambda\) is a hyperparameter balancing the two tasks. This method allows the model to learn dependencies between sentences more effectively.

\subsection{InLegalBERT Variants}
We experimented with different configurations of the InLegalBERT~\citet{paul-2022-pretraining} model to improve performance. These configurations vary in terms of the number of sentences provided as input during training and inference:
\begin{itemize}
    \item InLegalBERT(i): The model is trained and tested using only the current sentence \( i \).
    \item InLegalBERT(i-1, i): The model is trained with the previous sentence \( i-1 \) and the current sentence \( i \).
    \item InLegalBERT(i-2, i-1, i): The model is trained using the previous two sentences \( i-2, i-1 \) and the current sentence \( i \).
    \item InLegalBERT(i-1, i, i+1): The model is trained with the previous sentence, current sentence, and the next sentence.
\end{itemize}

\subsection{Incorporate Previous Sentence and Label}
We further explored methods where we provide the model with additional contextual information. In one variant, we concatenate the current sentence with the previous sentence and the true label of the previous sentence during training. This approach allows the model to leverage contextual information from preceding sentences to make better predictions. Another variant replaces the true label with the predicted label of the previous sentence during inference, simulating real-world conditions where true labels are unavailable. This method helps the model handle prediction errors and learn sequential dependencies between rhetorical roles.

\subsection{Self-Supervised Pre-Training with Role-Aware Transformers}

We propose a novel Role-Aware Transformer, which extends the standard transformer architecture by integrating role embeddings to represent rhetorical roles such as Facts, Issues, Arguments, and Reasoning. The model is pre-trained in a self-supervised manner on a large corpus of legal documents, allowing it to learn structural and contextual dependencies in legal discourse.

During pre-training, the model predicts masked tokens while leveraging sentence-level role embeddings. Unlike standard transformers, which process sentences without explicit role awareness, our approach incorporates additional role-specific information into the input embeddings. Specifically, each token embedding is enriched with a learned role embedding that represents its rhetorical role, allowing the model to develop a deeper understanding of legal text organization. This enhances the ability to distinguish between similar rhetorical roles and improves overall classification performance.

For pre-training, we initialize the model with InLegalBERT, a transformer specifically pre-trained on Indian legal documents. By incorporating rhetorical role awareness, this method enables the model to better capture the discourse structure of legal texts, leading to more accurate and context-aware classification outcomes.



\subsection{GNN with Document Context}
To capture the structural relationships between sentences, we propose a method that leverages Graph Neural Networks (GNNs). In this approach, each sentence in a document is represented as a node in a graph, and the edges between nodes are based on sentence order or semantic similarity. Sentence embedding generated via InLegalBERT, a pre-trained language model on the Indian legal domain, serves as a node feature. The GNN processes the graph by propagating information between connected sentences, allowing the model to capture both local and global contextual dependencies. The GNN processes this graph, allowing for information propagation and aggregation across connected sentences, which enhances understanding of interdependencies between sentences.

\subsection{\texttt{RhetoricLLaMA}}
To leverage the power of LLMs for rhetorical role prediction, we implemented \texttt{RhetoricLLaMA}, an instruction-tuned model based on LLaMA-2-7B \citet{touvron2023llama}. For this specific task, we fine-tuned the LLaMA-2-7B model on our \texttt{LegalSeg} dataset using instruction-tuning, a method designed to guide the model’s understanding of specific tasks through a set of structured instructions.

To enhance the model's ability to segment legal documents accurately, we developed a set of 16 instruction sets tailored to the nature of rhetorical role classification in legal texts. These instructions provided the model with explicit guidance on how to handle the different rhetorical roles in a legal document. A complete list of these instruction sets can be found in Table~\ref{tab:instruction_sets} in the Appendix.

\section{Evaluation Metrics}
To evaluate the performance of models, we adopt a set of standard metrics commonly used in classification tasks. For each sentence in the dataset, the predicted label (rhetorical role) is considered correct if it matches the label assigned by the human expert annotator.

We utilize macro-averaged Precision, Recall, F-score, Accuracy, and Matthew Correlation Coefficient (MCC) \citet{chicco2020advantages} as our primary evaluation metrics. Macro-averaging involves calculating these metrics for each class separately and then taking their average. This method is particularly beneficial as it prevents bias towards high-frequency classes, ensuring that all rhetorical roles are treated equally in the evaluation process.

\section{Results and Analysis}
\begin{table}[t]
\centering
\resizebox{\linewidth}{!}{%
\begin{tabular}{lccccc}
\toprule
\textbf{Model}                  & \textbf{Precision} & \textbf{Recall} & \textbf{F1-Score} & \textbf{Accuracy} & \textbf{MCC} \\ 
\midrule
MTL                             & 0.59               & 0.40            & 0.37              & 0.41              & \textbf{0.78}            \\ 
GNN                             & 0.64               & 0.50            & 0.54              & \textbf{0.64}              & 0.40         \\ 
Role-Aware                      & 0.21               & 0.20            & 0.14              & 0.50              & 0.04            \\ 
ToInLegalBERT                   & 0.67               & 0.60            & 0.62              & \textbf{0.64}              & 0.52         \\ 
LLaMA-2 (Quantized)   & 0.17                  & 0.16               & 0.09                 & 0.20                 & 0.3            \\ 
LLaMA-2 (Unquantized)   & 0.19                  & 0.15               & 0.08                 & 0.25                 & 0.05            \\
\texttt{RhetoricLLaMA}                   & 0.19               & 0.15            & 0.09              & 0.39              & 0.02         \\ 
InLegalBERT(i)                  & 0.57               & 0.45            & 0.49              & 0.53              & 0.45         \\ 
InLegalBERT(i-1, i)             & 0.60               & 0.53            & 0.55              & 0.57              & 0.50         \\ 
InLegalBERT(i-2, i-1, i)        & 0.62               & 0.56            & 0.58              & 0.59              & 0.52         \\ 
InLegalBERT(i-1, i, i+1)        & 0.61               & 0.56            & 0.58              & 0.59              & 0.52         \\ 
InLegalBERT(i-1, label\_t, i)   & 0.63                  & 0.32               & 0.34                 & 0.45                 & 0.22            \\ 
InLegalBERT(i-1, label\_p, i)   & 0.54               & 0.46            & 0.48              & 0.52              & 0.35         \\ 
Hier\_BiLSTM CRF                & \textbf{0.78}               & \textbf{0.77}            & \textbf{0.77}              & 0.62              & 0.68         \\ 

\bottomrule
\end{tabular}
}
\caption{Performance Comparison of Models on Rhetorical Role Classification. In the Model column, \(i\) indicates the current sentence, \(i-1\) means the previous sentence, and \(i+1\) means the next sentence. \texttt{label\_t} and \texttt{label\_p} refer to the true and predicted labels of the previous sentences. The best results are in bold.}
\label{tab:model_performance}
\end{table}

In this section, we present the results of our experiments on rhetorical role classification and analyze the performance of different models. Table \ref{tab:model_performance} summarizes the evaluation metrics for each model.

\subsection{Model Performance}
Among the evaluated models, the hierarchical BiLSTM-CRF achieves the highest overall performance. The sequential nature of BiLSTM allows the model to capture dependencies between sentences, while the CRF layer explicitly models label transitions, refining predictions by enforcing structural coherence. This ability to learn the transition relationships between rhetorical roles plays a crucial role in classification, as labels in legal documents follow a structured sequence. For example, an issue is likely to be followed by supporting arguments and eventually a decision. The ability to maintain coherence in predictions by capturing dependencies between consecutive sentences makes the BiLSTM-CRF model more effective in comparison to models that classify each sentence independently. Prior studies in structured text classification have similarly observed the benefits of explicit modeling of transition relationships between labels, as seen in \citet{bhattacharya2019identification, modi-etal-2023-semeval, santosh2024hiculr}.

In contrast, transformer-based models such as ToInLegalBERT, InLegalBERT, and Role-Aware Transformers process sentences independently, limiting their ability to model long-range dependencies within legal documents. These models rely primarily on self-attention mechanisms, which work well for general NLP tasks but struggle to capture structured rhetorical transitions without explicit sequential modeling. ToInLegalBERT, which integrates sentence-level positional encodings and hierarchical structuring, performs better than standard BERT-based models, highlighting the benefit of incorporating document structure into transformers.

The Graph Neural Network model performs competitively by effectively propagating contextual information across sentence nodes, capturing both local and global dependencies within legal documents. Among the InLegalBERT variants, the model trained using the current sentence along with two preceding sentences achieves the best performance, reinforcing the importance of sentence context in improving classification accuracy.

The Multi-Task Learning model, which incorporates label shift prediction as an auxiliary task, achieves moderate performance. While this method aims to capture role transitions, the additional complexity may have introduced challenges in optimization. Despite this, multitask learning remains a promising approach, particularly when combined with stronger baseline models.

The \texttt{RhetoricLLaMA} model, despite being instruction-tuned, did not perform as strongly as expected. While large language models like LLaMA-2-7B have achieved success in NLP, their effectiveness in specialized tasks such as rhetorical role classification remains limited without extensive domain-specific fine-tuning. Further research is needed to optimize large language models for structured legal NLP tasks.

\subsection{Impact of Transition Relationships in Classification}
Our experiments highlight the critical role of transition relationships between rhetorical roles in improving classification performance. Models such as the BiLSTM-CRF explicitly model these transitions, allowing them to maintain coherence in predictions by capturing dependencies between consecutive sentences. This is particularly advantageous because legal documents are highly structured, with rhetorical roles appearing in predictable sequences. In contrast, models that classify each sentence in isolation struggle to maintain contextual consistency, leading to higher misclassification rates.

For instance, when a sentence is labeled as an issue, the subsequent sentences are highly likely to be arguments or facts rather than a decision. CRF layers enforce these structural constraints, making BiLSTM-CRF more effective than independent sentence classifiers. This aligns with previous findings in rhetorical role classification, where modeling dependencies between sequential labels significantly improved performance in structured text classification tasks.

\subsection{Justification for Predicted Labels Showing Higher Performance}
An interesting observation from Table \ref{tab:model_performance} is that models using predicted labels for previous sentences sometimes outperform those using true labels. This initially appears counterintuitive, but a plausible explanation is that during training, both true labels and predicted labels were provided to the model, allowing it to learn effective dependencies. However, during testing, true labels are not available, meaning models trained exclusively with true labels may not learn to handle missing labels during inference. In contrast, models using predicted labels during training are already exposed to prediction noise, making them better adapted to real-world inference conditions where true labels are not available.

This suggests that training models to rely on predicted labels during both training and inference improves robustness, as the model learns to correct potential errors in label predictions over multiple steps. However, further research is needed to analyze whether explicitly modeling label uncertainty could further enhance performance.

\subsection{Impact of Instruction-Tuning in \texttt{RhetoricLLaMA}}
We conducted extensive experiments to analyze the impact of instruction-tuning in \texttt{RhetoricLLaMA} by comparing it against Vanilla LLaMA models in both quantized (4-bits) and unquantized forms. Despite leveraging large-scale pre-trained models, the instruction-tuned \texttt{RhetoricLLaMA} did not achieve the expected performance, suggesting that rhetorical role classification in legal texts requires more specialized adaptations.

The comparison revealed that the instruction-tuned model performed slightly better than the Vanilla LLaMA model but still lagged behind traditional transformer-based models like \texttt{ToInLegalBERT} and BiLSTM-CRF. While instruction-tuning provides explicit task-specific guidance, our results indicate that for highly specialized domains such as legal NLP, additional domain-specific pre-training and refined instruction sets are necessary to enhance model performance.

\subsection{Error Analysis}
\begin{figure}[t]
    \centering
    \includegraphics[width=\linewidth]{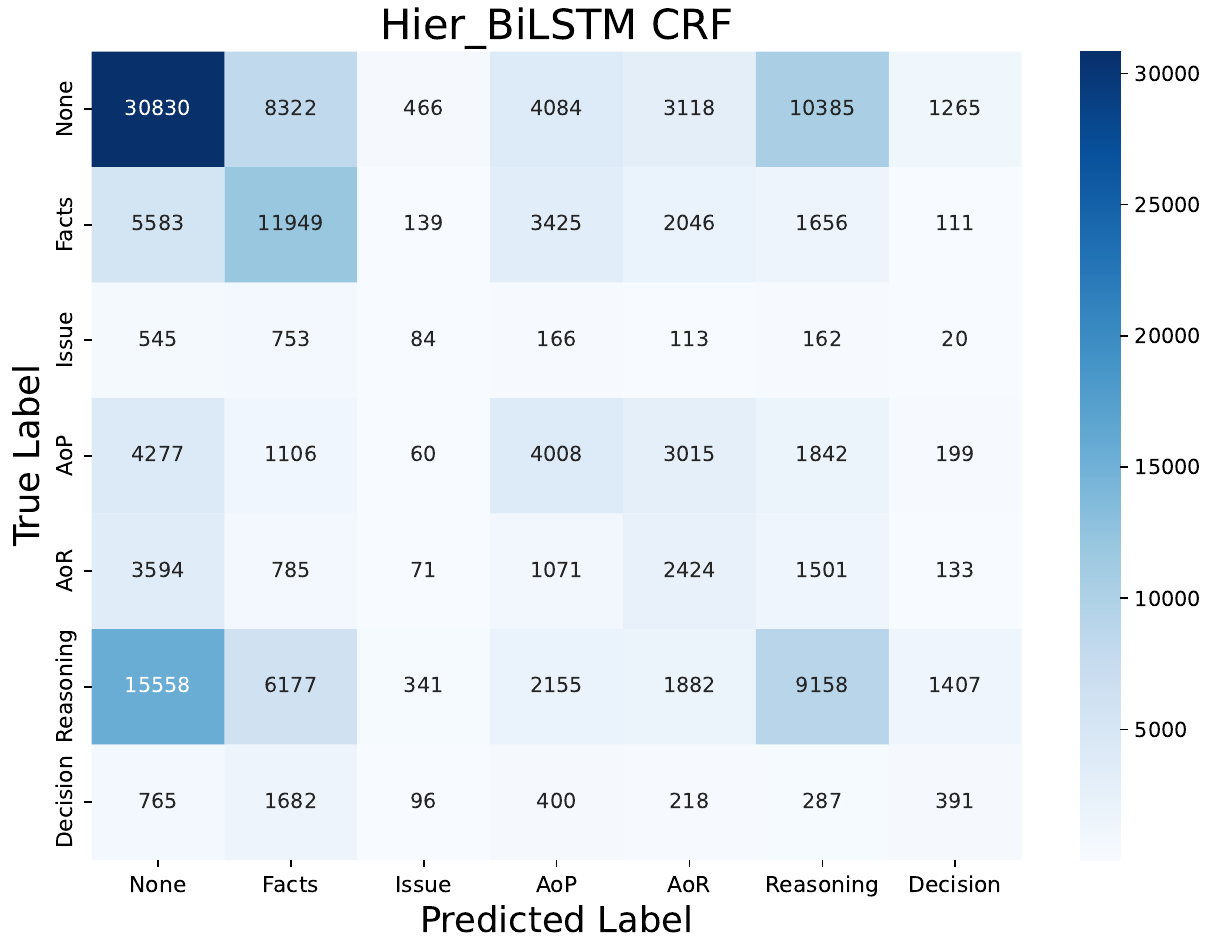}
    \caption{Confusion matrix for rhetorical role classification using Hierarchical BiLSTM-CRF model.}
    \label{fig:hier_bilstm_crf}
\end{figure}

Our error analysis revealed that the models struggled primarily with distinguishing between closely related rhetorical roles, such as Facts and Reasoning, due to the overlap in their language and structure within legal documents. This challenge is clearly illustrated in the confusion matrix of the Hierarchical BiLSTM-CRF model Figure~\ref{fig:hier_bilstm_crf}, which shows frequent misclassifications between these roles. Similarly, confusion between Arguments of Petitioner and Arguments of Respondent was prevalent, as both often exhibit similar language patterns, further complicating accurate classification. Models that incorporated contextual information from preceding or following sentences demonstrated some improvement in reducing these errors, particularly for roles requiring a clear transition, such as Issue and Decision. However, despite this improvement, the context-aware models still encountered difficulties, suggesting that the rhetorical role boundaries within these transitions are not always well-defined. Another critical issue identified was class imbalance. More frequent labels like None and Facts were consistently overpredicted, leading to lower precision for less frequent labels such as Issue and Decision. This imbalance skewed the performance, resulting in models favoring high-frequency roles at the expense of accuracy for underrepresented roles.
Figures~\ref{fig:mtl},~\ref{fig:gnn},~\ref{fig:toinlegalbert},~\ref{fig:rhetoricllama},~\ref{fig:inlegalbert_i},~\ref{fig:inlegalbert_i-1_i},~\ref{fig:inlegalbert_i-2_i-1_i},~\ref{fig:inlegalbert_i-1_i_i+1},~\ref{fig:inlegalbert_label_t},~\ref{fig:inlegalbert_label_p} illustrating the confusion matrices for other models, are provided in the Appendix due to space constraints. These figures further highlight the patterns of misclassification and the impact of various model architectures on error distribution. Addressing these issues, particularly through improved handling of context, mitigating class imbalance, and minimizing the propagation of sequential errors, remains a critical area for future research and model refinement.


\section{Conclusion and Future Work}
In this work, we addressed the challenging task of rhetorical role classification in legal documents by introducing the \texttt{LegalSeg} dataset, the largest annotated dataset for this task. \texttt{LegalSeg}, provides a significant resource for advancing research in this domain. We evaluated multiple models, including \texttt{RhetoricLLaMA}, ToInLegalBERT, Role-aware, and GNNs. Our results show that models incorporating both sequential and contextual information, such as Hierarchical BiLSTM-CRF and ToInLegalBERT, perform best in identifying and classifying rhetorical roles in legal texts. We also demonstrated that adding sentence-level context improves the model's ability to capture transitions between rhetorical roles, reducing errors caused by the inherent similarity between roles like Facts and Reasoning.

Despite these advancements, our error analysis revealed several challenges, such as misclassification between similar roles and the cascading effect of label prediction errors. Furthermore, class imbalance remains a significant issue, with frequent misclassifications of minority labels.

For future work, we aim to explore more sophisticated techniques to handle class imbalance, such as advanced sampling strategies and loss function adjustments. Additionally, refining models' ability to capture long-range dependencies and leveraging more robust pre-training strategies could further enhance the performance of LLMs. We also plan to incorporate more domain-specific knowledge into the models and experiment with cross-domain transfer learning to improve their adaptability across different legal contexts. 

\section*{Acknowledgements}
We would like to express our gratitude to the anonymous reviewers for their insightful comments and constructive feedback, which have significantly improved the quality of this work. We also sincerely thank the student research assistants from various law colleges for their invaluable contributions in annotating the documents. Their efforts have been instrumental in the development of this research.  

This work was supported by the ``Research-I Foundation'' at the Dept. of Computer Science and Engineering, IIT Kanpur, which has generously funded the conference travel.

\section*{Limitations}
While this study makes significant strides in rhetorical role classification for legal documents, a few areas remain where further refinement could enhance the approach. These areas are opportunities for future work rather than major limitations and are not expected to diminish the contribution of this research.

The \texttt{LegalSeg} dataset, while being the largest and most comprehensive of its kind for Indian legal judgments, is understandably specialized in the context of the Indian judiciary system. This focus provides unique insights into this particular legal domain. However, it is acknowledged that the models may require adaptation to handle legal documents from different jurisdictions. This does not limit the validity of our findings but opens a path for future research into cross-jurisdictional generalization using transfer learning techniques or domain adaptation strategies, which are common challenges in domain-specific NLP.

The class imbalance in the dataset, which is inherent in most real-world legal corpora, reflects the natural distribution of rhetorical roles in judgments. While some roles like Issue and Decision are less frequent, this mirrors their actual occurrence in legal texts. We have taken steps to mitigate this issue through advanced modeling techniques such as label shift prediction and the incorporation of contextual information. Future work could explore further enhancements, such as data augmentation or more refined class-weighting techniques, to boost performance on the less frequent roles.

Additionally, the computational requirements of models like ToInLegalBERT and \texttt{RhetoricLLaMA} are justified given the complexity and the high accuracy they provide. These models are aligned with state-of-the-art practices in NLP, which involve significant computational demands. While this may pose a challenge for deployment in low-resource environments, it is important to note that high-performance models are typically developed on powerful infrastructures and then optimized for more practical use cases through techniques such as model pruning, quantization, or distillation, which can be addressed in future work.

The overlap in rhetorical roles, such as between Facts and Reasoning, is an inherent challenge in legal discourse due to the intertwined nature of legal arguments and fact presentation. The models already handle these overlaps competently, and our use of sequential and contextual information improves performance. However, we recognize that future refinements, such as more sophisticated context-aware mechanisms or hybrid models that integrate symbolic reasoning with machine learning, could offer even greater differentiation between closely related roles.

In conclusion, the challenges discussed here are not insurmountable and represent common issues in the evolving field of legal NLP. This work provides a strong foundation for addressing these aspects, and we are confident that the solutions proposed will inspire future innovations and improvements. This manuscript significantly advances the state of the art in rhetorical role classification, and any remaining opportunities for refinement will only serve to further enhance the impact of this research.

\section*{Ethics Statement}
This research was conducted with a strong commitment to ethical standards. The \texttt{LegalSeg} dataset comprises publicly available Indian legal judgments, ensuring no private or sensitive data was included. Anonymization was applied where necessary, and all data was collected in compliance with legal and privacy regulations.

In the process of annotating the dataset, law students from various institutions were involved. These annotators were treated with fairness and respect, and they were appropriately compensated for their contributions. Informed consent was obtained from all participants before their involvement in the project. While the students provided essential support in developing the dataset, they are not listed as co-authors of this manuscript to maintain the academic integrity of the publication. Instead, their valuable contributions are acknowledged separately.

The models developed are designed to support, not replace, legal professionals. We advocate for their responsible use, emphasizing the need for human oversight when applied in real-world legal contexts. This research adheres to ethical guidelines in authorship, data handling, and participant involvement, ensuring that all contributions are treated with fairness and respect.
\newpage

\bibliography{custom}

\newpage
\appendix
\newpage
\section{Experimental Setup and Hyper-parameters}
We conducted experiments across several models, utilizing different architectures and training techniques tailored to rhetorical role classification tasks. Below, we provide an overview of the key experimental setups and hyper-parameters used.

\subsection{\texttt{RhetoricLLaMA} Training Procedure}
\texttt{RhetoricLLaMA}, built on the LLaMA-2-7B model, was fine-tuned with Bfloat16 precision using a single A100 GPU with 40GB memory. Given the computational constraints, the model was optimized for efficiency, with training lasting 48 hours. A maximum token length of 1000 was used, and Low-Rank Adaptation (LoRA) was employed with a rank of 16, alpha set to 64, and a dropout rate of 0.1. The model leveraged flash-attention 2 for faster training. We applied a Paged Adam optimizer with a learning rate of 1e-4 and a cosine learning rate scheduler, along with gradient accumulation steps of 4. The model trained for 52,617 steps, corresponding to 3 epochs.

\subsection{Transformers Training Hyper-parameters}
For the Role-Aware Transformers, built upon the InLegalBERT model, pre-training involved self-supervised tasks such as Masked Language Modeling (MLM) with role embeddings added. The model processed a maximum sequence length of 512 tokens with a batch size of 4, running for 20 epochs. The learning rate was set to 2e-5, using the AdamW optimizer. Class weights were applied to handle the imbalance in rhetorical roles, and early stopping was used to prevent overfitting.

\subsection{Graph Neural Networks (GNN) with Document Context}
We utilized a Graph Neural Network (GNN) architecture to model sentence relationships within legal documents. A two-layer Graph Convolutional Network (GCN) processed sentence embeddings from InLegalBERT. The first and second GCN layers both had output dimensions of 128, using ReLU activations. The model was trained for 10 epochs with a learning rate of 1e-4 and employed a Cross-Entropy Loss function. Graphs were constructed with edges between consecutive sentences, capturing both sequential and semantic relationships.

\subsection{Incorporating Previous Sentence and Actual Label}
In this method, the input to the model combined the current sentence with the previous sentence and its actual rhetorical role label. The model used InLegalBERT with a maximum sequence length of 512 tokens, trained for 5 epochs with a learning rate of 2e-5. This approach provided explicit sequential context and utilized Cross-Entropy Loss with class weights to manage class imbalance.

\subsection{Incorporating Previous Sentence and Predicted Label}
Extending the previous approach, this variant incorporated the predicted label of the previous sentence, simulating real-world conditions. The same configuration was used as in the previous model, but the predicted label replaced the actual label during both training and inference.

\subsection{Common Settings Across Models}
All models were evaluated using Accuracy, Precision, Recall, F1 Score, and Matthews Correlation Coefficient (MCC). The experiments utilized PyTorch and Hugging Face Transformers libraries, with PyTorch Geometric handling the graph data in the GNN method. All models were trained on machines with NVIDIA GPUs for parallel computation.

\begin{table*}[h!]
\centering
\resizebox{\textwidth}{!}{%
\begin{tabular}{|l|l|}
\hline
\textbf{Label}                   & \multicolumn{1}{|c|}{\textbf{{Sentences}}}
\\ \hline
Fact                    & \begin{tabular}[c]{@{}l@{}}For the sake of convenience, we are referring to the facts of Civil Appeal No.1328 of 2021.\end{tabular}                                              \\ \hline
Fact                    & \begin{tabular}[c]{@{}l@{}}At the time of the assessment proceedings, the Assessee submitted a revised computation \\of income by revising its claim of deduction under Section 80IA of the Act.\end{tabular}                                                                                                                                                   \\ \hline
Issue   & \begin{tabular}[c]{@{}l@{}}The Income Tax Appellate Tribunal (hereinafter the Tribunal), upheld the decision of the \\Appellate Authority on the issue of deduction under Section 80IA.\end{tabular}                                                                                                                                                          \\ \hline
Issue   & \begin{tabular}[c]{@{}l@{}}The High Court refused to interfere with the Tribunals order as far as the issue on deduction \\under Section 80IA is concerned.\end{tabular}                                                                                      \\ \hline
\begin{tabular}[c]{@{}l@{}}Arguments of \\Petitioner (AoP)  \end{tabular}              & \begin{tabular}[c]{@{}l@{}}Mr. Arijit Prasad, learned Senior Counsel appearing on behalf of the Revenue, submitted \\that Section 80AB of the Act contemplates deductions in respect of incomes against income \\of the nature specified in the relevant section.\end{tabular}                                                        \\ \hline
\begin{tabular}[c]{@{}l@{}}Arguments of \\Petitioner (AoP)  \end{tabular}                 & \begin{tabular}[c]{@{}l@{}}According to him, the phrase derived from in subsection (1) of Section 80IA of the Act \\indicates that the computation of deduction is restricted only to the profits and gains from the \\eligible business.\end{tabular} \\ \hline

\begin{tabular}[c]{@{}l@{}}Arguments of \\Respondent (AoR)  \end{tabular}                & \begin{tabular}[c]{@{}l@{}}In response, the Assessee supported the order passed by the Appellate Authority which was \\upheld by the Tribunal and the High Court.\end{tabular}                                                                                                                        \\ \hline

\begin{tabular}[c]{@{}l@{}}Arguments of \\Respondent (AoR)  \end{tabular}                 & \begin{tabular}[c]{@{}l@{}}He submitted that there is no indication in subsection (5) of Section 80IA that the deduction \\under subsection (1) is restricted to business income only.\end{tabular}                                                                                                                             \\ \hline

Reasoning  & \begin{tabular}[c]{@{}l@{}}As stated above, Section 80AB was inserted in the year 1981 to get over a judgment of this \\Court in Cloth Traders (P) Ltd. (supra).\end{tabular}                                                                                                     \\ \hline
Reasoning   & \begin{tabular}[c]{@{}l@{}}On the question of existence of vacancies, although learned counsel for the appellant \\submitted that vacancies are still lying there, which submission however has been refuted by \\the learned counsel for the State of Rajasthan.\end{tabular}                                                                                                                                                       \\ \hline
Decision & \begin{tabular}[c]{@{}l@{}}For the aforementioned reasons, the Appeal is dismissed qua the issue of the extent of \\deduction under Section 80IA of the Act.\end{tabular}                                                                                        \\ \hline
Decision & \begin{tabular}[c]{@{}l@{}}The assets of the Corporate Debtor shall be managed strictly in terms of the provisions of \\the IBC.\end{tabular}                                                                                                                                                                              \\ \hline
None                    & Clause 11(b) reads as follows 11.                                                                                                                                                                                                                                                                                                                 \\ \hline
None                    & The clause reads thus 12 Miscellaneous.                                                                                                                                                                                                                                                                                  \\ \hline

\end{tabular}
}
\caption{Example sentences for each label.}
\label{tab:rr-examples}
\end{table*}

\begin{table*}[h]
\centering
\resizebox{\textwidth}{!}{%
\begin{tabular}{|c|l|}
\hline
\multicolumn{2}{|c|}{\textbf{\textcolor{blue}{Instruction Sets}}} \\ \hline

\multicolumn{1}{|c|}{1}  &  \begin{tabular}[c]{@{}l@{}}Analyze the given legal sentence and predict its rhetorical role as a number: None-0, Facts-1, Issue-2, \\Arguments of Petitioner-3, Arguments of Respondent-4, Reasoning-5, Decision-6.\end{tabular} \\ \hline

\multicolumn{1}{|c|}{2}  &  \begin{tabular}[c]{@{}l@{}}Determine the rhetorical function of this sentence from a court case and provide its corresponding number: \\None-0, Facts-1, Issue-2, Arguments of Petitioner-3, Arguments of Respondent-4, Reasoning-5, Decision-6.\end{tabular} \\ \hline

\multicolumn{1}{|c|}{3}  &  \begin{tabular}[c]{@{}l@{}}Based on the content of the following legal text, classify its rhetorical role by selecting the appropriate number: \\None-0, Facts-1, Issue-2, Arguments of Petitioner-3, Arguments of Respondent-4, Reasoning-5, Decision-6.\end{tabular} \\ \hline

\multicolumn{1}{|c|}{4}  &  \begin{tabular}[c]{@{}l@{}}Identify the rhetorical category of this legal statement and provide the number that represents it: None-0, \\Facts-1, Issue-2, Arguments of Petitioner-3, Arguments of Respondent-4, Reasoning-5, Decision-6.\end{tabular} \\ \hline

\multicolumn{1}{|c|}{5}  &  \begin{tabular}[c]{@{}l@{}}Evaluate the rhetorical purpose of the provided legal sentence and label it with the correct number: None-0, \\Facts-1, Issue-2, Arguments of Petitioner-3, Arguments of Respondent-4, Reasoning-5, Decision-6.\end{tabular} \\ \hline

\multicolumn{1}{|c|}{6}  &  \begin{tabular}[c]{@{}l@{}}Assign a number to the rhetorical role of this sentence from a legal case, choosing from: None-0, Facts-1, \\Issue-2, Arguments of Petitioner-3, Arguments of Respondent-4, Reasoning-5, Decision-6.\end{tabular} \\ \hline

\multicolumn{1}{|c|}{7}  &  \begin{tabular}[c]{@{}l@{}}Review the legal statement and predict its rhetorical function using the corresponding number: None-0, \\Facts-1, Issue-2, Arguments of Petitioner-3, Arguments of Respondent-4, Reasoning-5, Decision-6.\end{tabular} \\ \hline

\multicolumn{1}{|c|}{8}  &  \begin{tabular}[c]{@{}l@{}}Examine this legal text and determine its rhetorical role by outputting the appropriate number: None-0, Facts-1, \\Issue-2, Arguments of Petitioner-3, Arguments of Respondent-4, Reasoning-5, Decision-6.\end{tabular} \\ \hline

\multicolumn{1}{|c|}{9}  &  \begin{tabular}[c]{@{}l@{}}Categorize the rhetorical purpose of the following sentence from a court proceeding with a number: None-0, \\Facts-1, Issue-2, Arguments of Petitioner-3, Arguments of Respondent-4, Reasoning-5, Decision-6.\end{tabular} \\ \hline

\multicolumn{1}{|c|}{10} &  \begin{tabular}[c]{@{}l@{}}Analyze the provided legal sentence and classify it into its rhetorical role, outputting only the number: None-0,\\ Facts-1, Issue-2, Arguments of Petitioner-3, Arguments of Respondent-4, Reasoning-5, Decision-6.\end{tabular} \\ \hline

\multicolumn{1}{|c|}{11} &  \begin{tabular}[c]{@{}l@{}}Determine the appropriate number for the rhetorical category of this legal text: None-0, Facts-1, Issue-2, \\Arguments of Petitioner-3, Arguments of Respondent-4, Reasoning-5, Decision-6.\end{tabular} \\ \hline

\multicolumn{1}{|c|}{12} &  \begin{tabular}[c]{@{}l@{}}Assign a numerical label to the rhetorical role of this statement in a legal case: None-0, Facts-1, Issue-2, \\Arguments of Petitioner-3, Arguments of Respondent-4, Reasoning-5, Decision-6.\end{tabular} \\ \hline

\multicolumn{1}{|c|}{13} &  \begin{tabular}[c]{@{}l@{}}Predict the number that corresponds to the rhetorical function of the following legal sentence: None-0, Facts-1, \\Issue-2, Arguments of Petitioner-3, Arguments of Respondent-4, Reasoning-5, Decision-6.\end{tabular} \\ \hline

\multicolumn{1}{|c|}{14} &  \begin{tabular}[c]{@{}l@{}}Identify the number that represents the rhetorical role of this legal text: None-0, Facts-1, Issue-2, Arguments of \\Petitioner-3, Arguments of Respondent-4, Reasoning-5, Decision-6.\end{tabular} \\ \hline

\multicolumn{1}{|c|}{15} &  \begin{tabular}[c]{@{}l@{}}Analyze this legal statement and assign the number that best matches its rhetorical function: None-0, Facts-1, \\Issue-2, Arguments of Petitioner-3, Arguments of Respondent-4, Reasoning-5, Decision-6.\end{tabular} \\ \hline

\multicolumn{1}{|c|}{16} &  \begin{tabular}[c]{@{}l@{}}Classify the following sentence from a court case by selecting its rhetorical role number: None-0, Facts-1, \\Issue-2, Arguments of Petitioner-3, Arguments of Respondent-4, Reasoning-5, Decision-6.\end{tabular} \\ \hline
\end{tabular}
}
\caption{Instruction Sets for Predicting the Roles}
\label{tab:instruction_sets}
\end{table*}


\begin{figure}[ht]
    \centering
    \includegraphics[width=\linewidth]{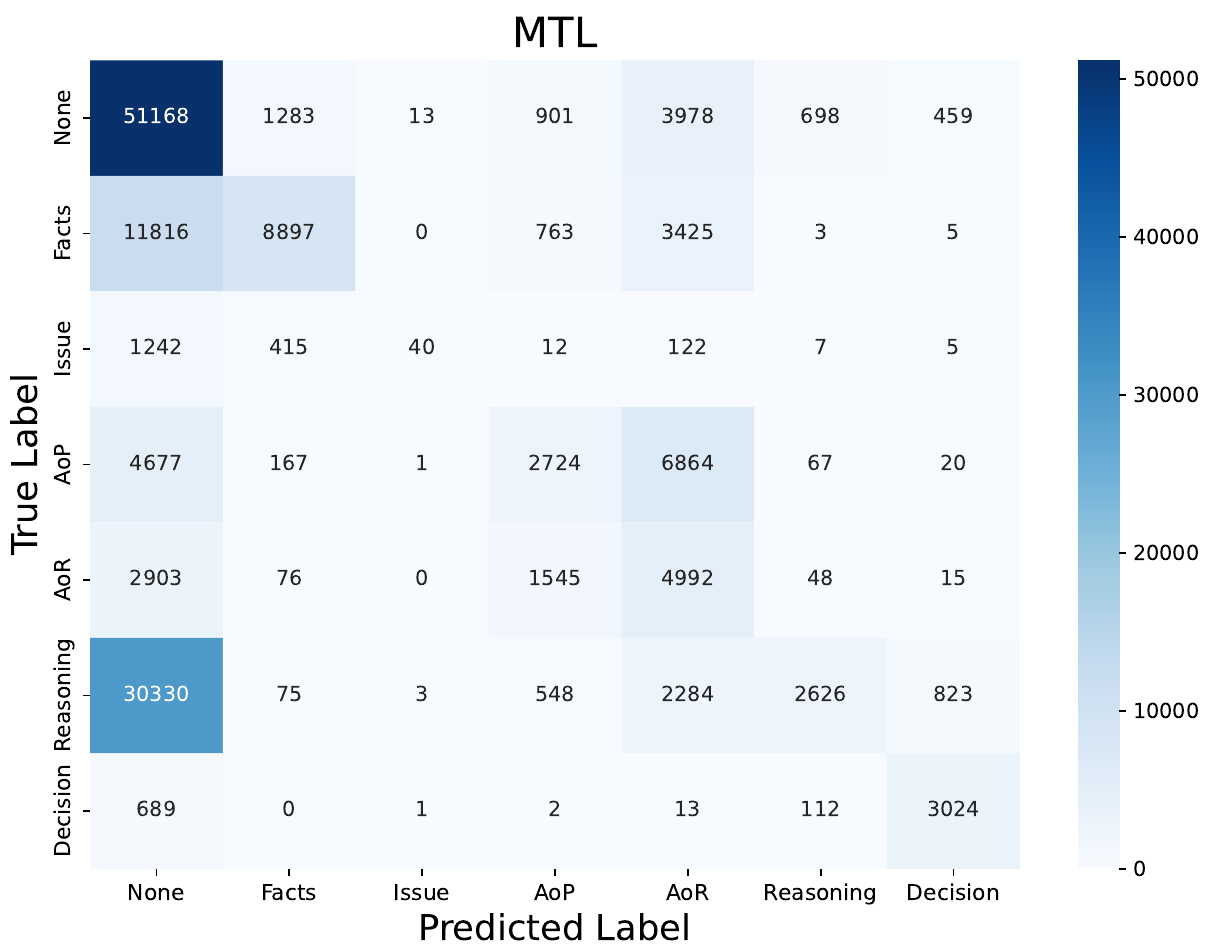}
    \caption{Confusion matrix for rhetorical role classification using the Multi-Task Learning (MTL) model.}
\label{fig:mtl}

\end{figure}

\begin{figure}[ht]
    \centering
    \includegraphics[width=\linewidth]{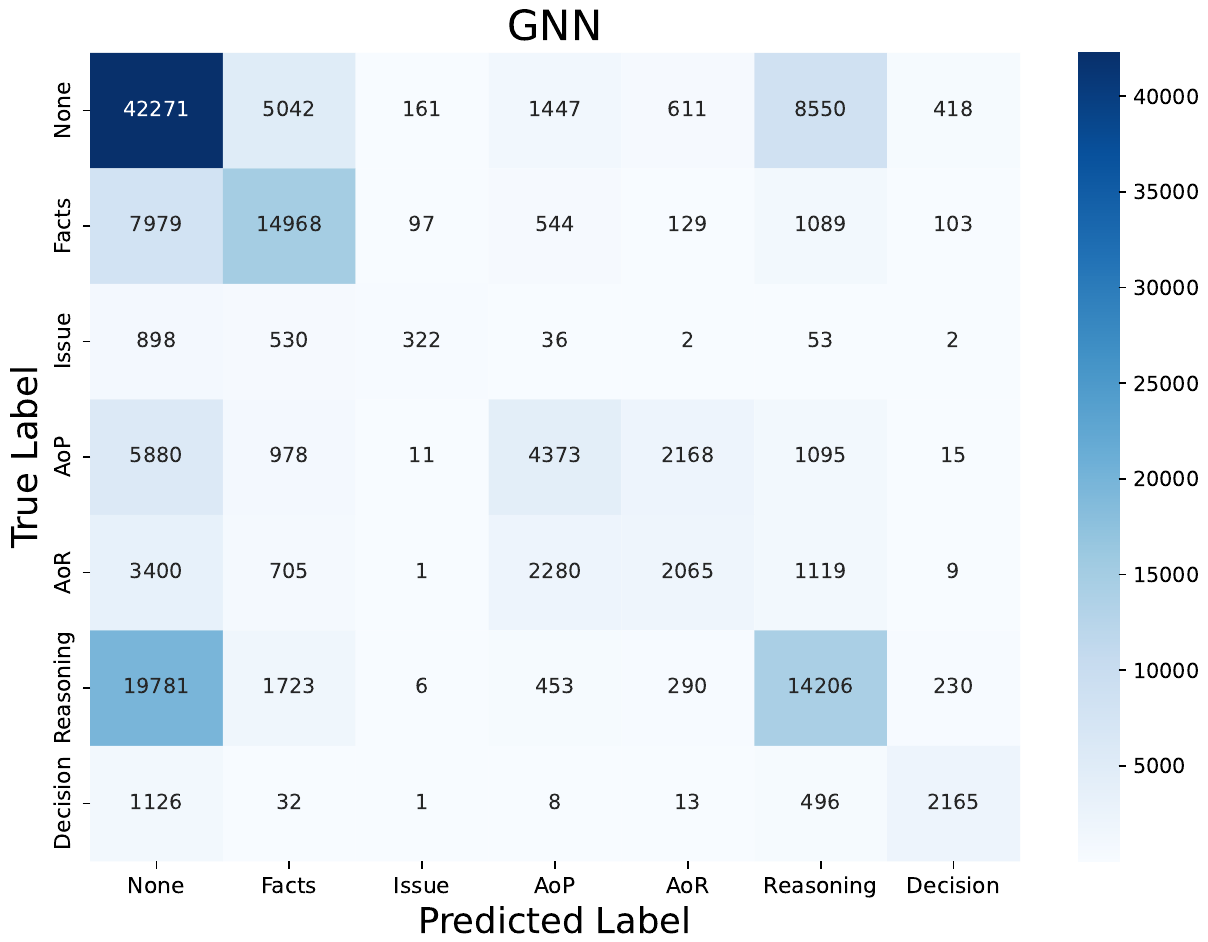}
    \caption{Confusion matrix for rhetorical role classification using GNN.}
\label{fig:gnn}
\end{figure}

\begin{figure}[ht]
    \centering
    \includegraphics[width=\linewidth]{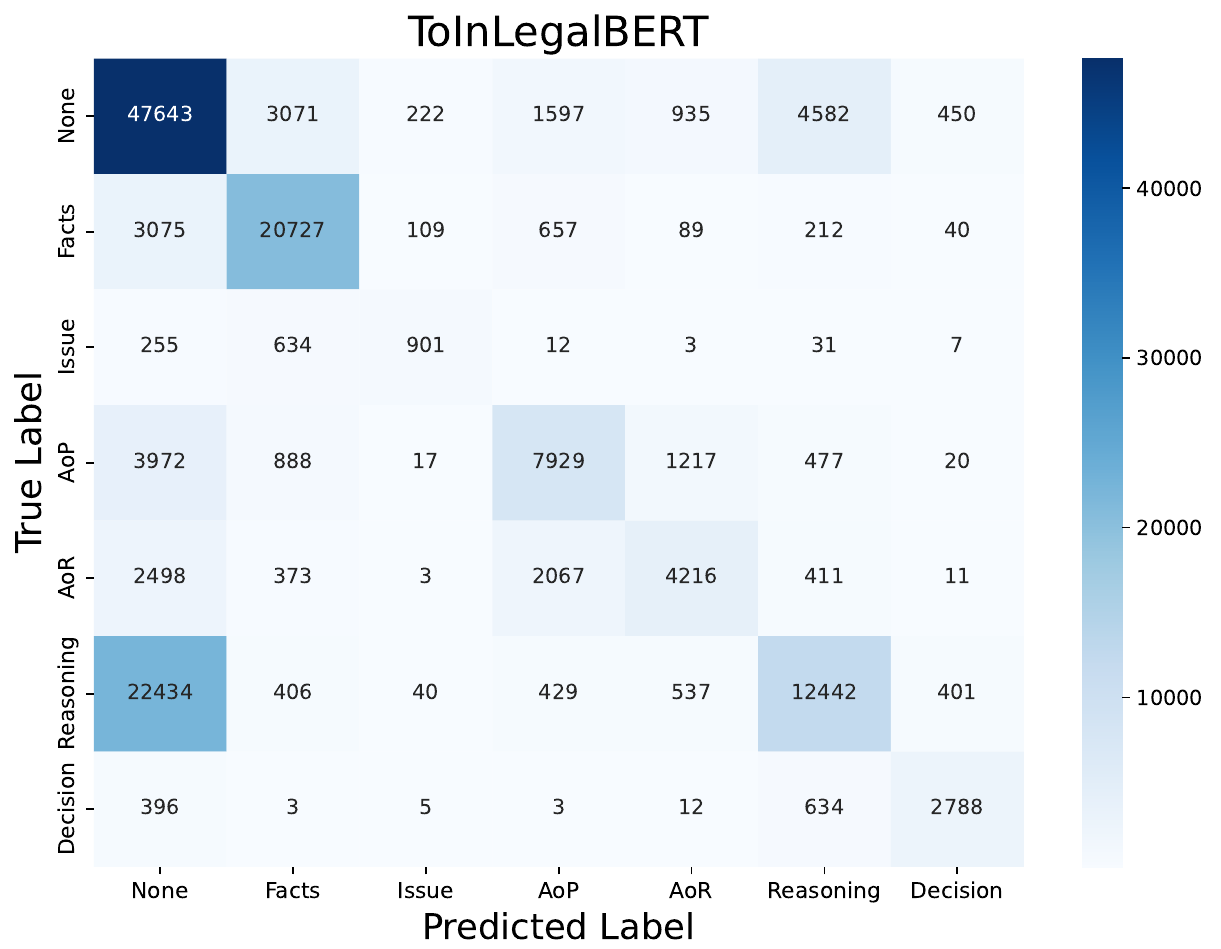}
    \caption{Confusion matrix for rhetorical role classification using TransformerOverInLegalBERT (ToInLegalBERT).}
\label{fig:toinlegalbert}
\end{figure}

\begin{figure}[ht]
    \centering
    \includegraphics[width=\linewidth]{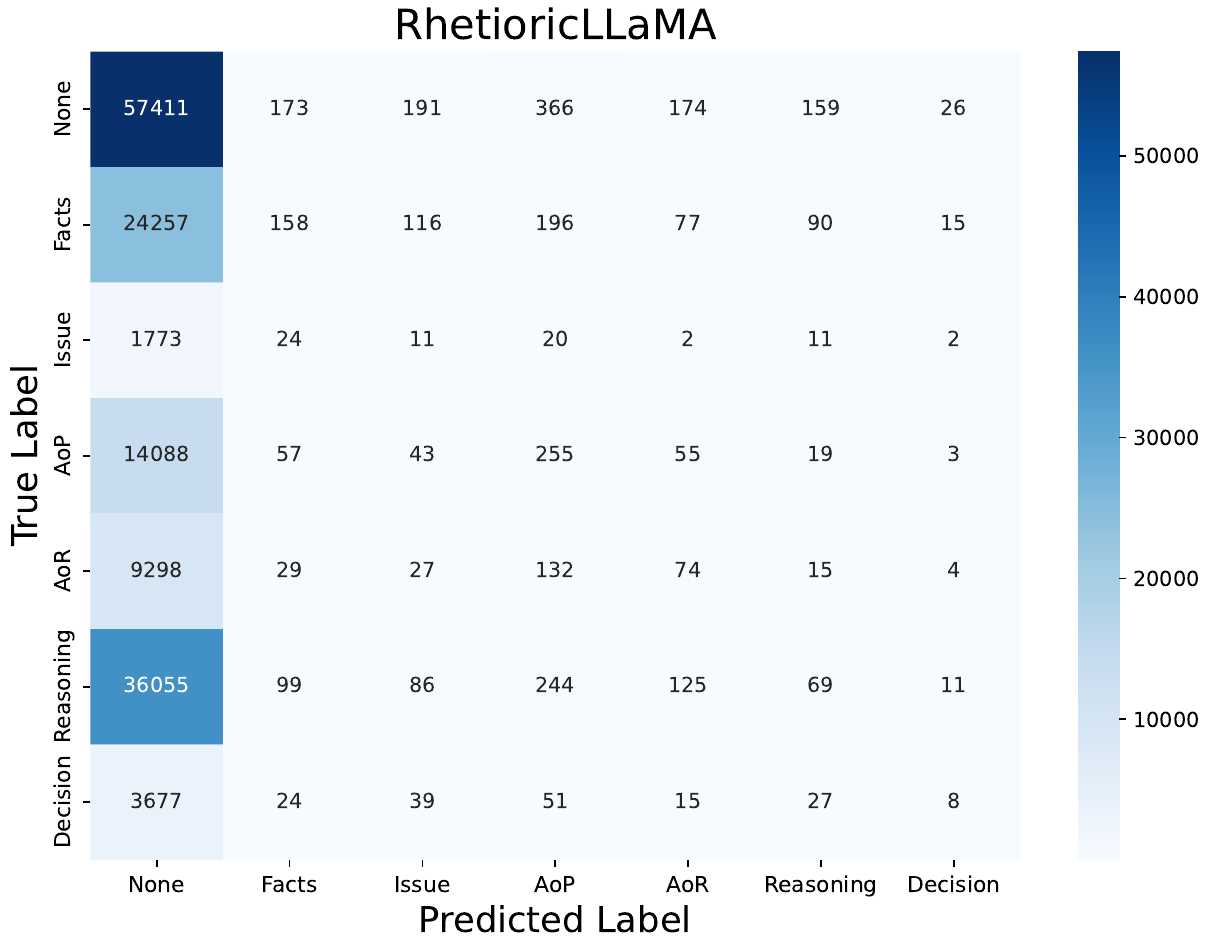}
    \caption{Confusion matrix for rhetorical role classification using \texttt{RhetoricLLaMA}, an instruction-tuned large language model.}
\label{fig:rhetoricllama}
\end{figure}

\begin{figure}[ht]
    \centering
    \includegraphics[width=\linewidth]{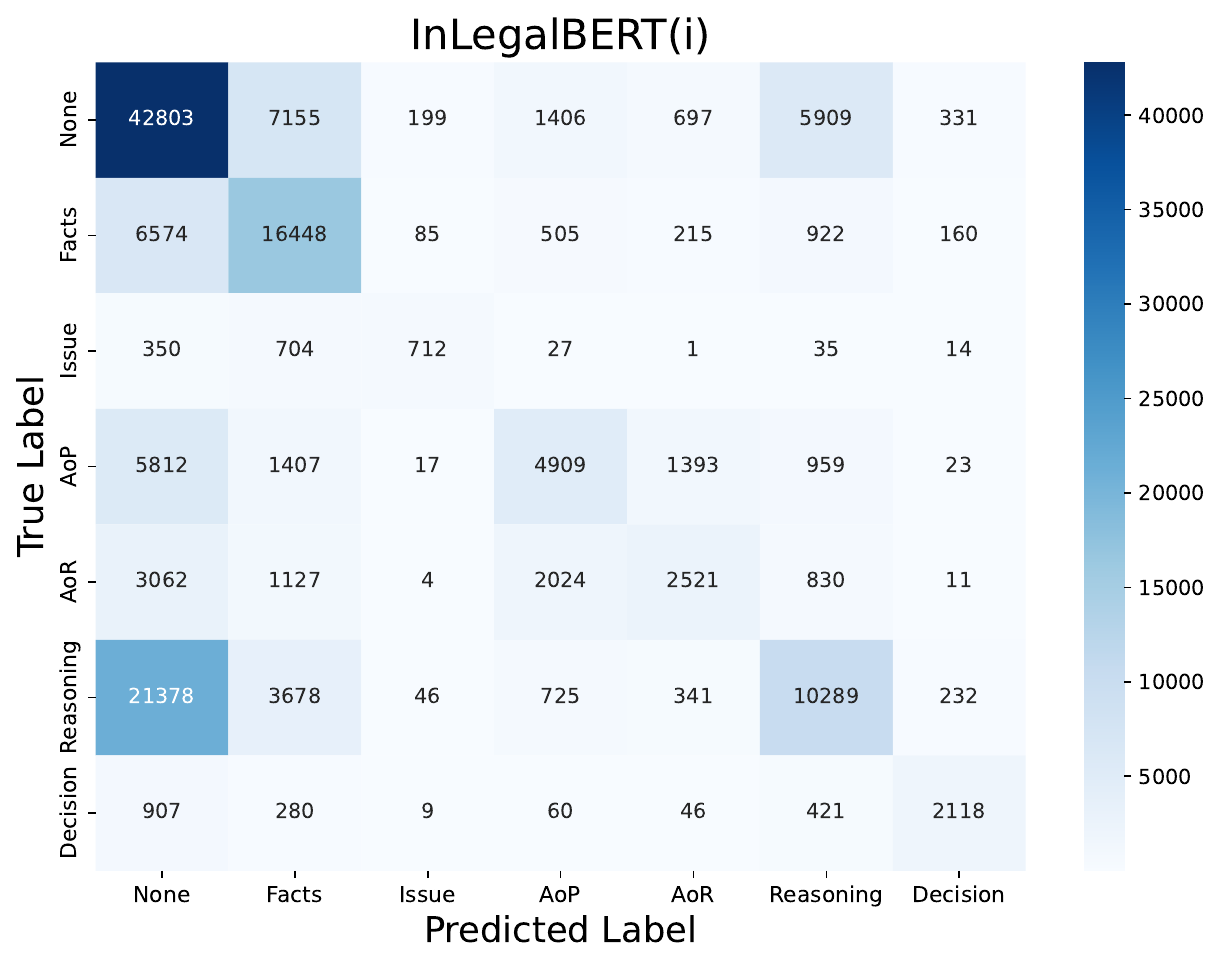}
    \caption{Confusion matrix for rhetorical role classification using InLegalBERT model with the current sentence (i) as input.}
    \label{fig:inlegalbert_i}
\end{figure}

\begin{figure}[ht]
    \centering
    \includegraphics[width=\linewidth]{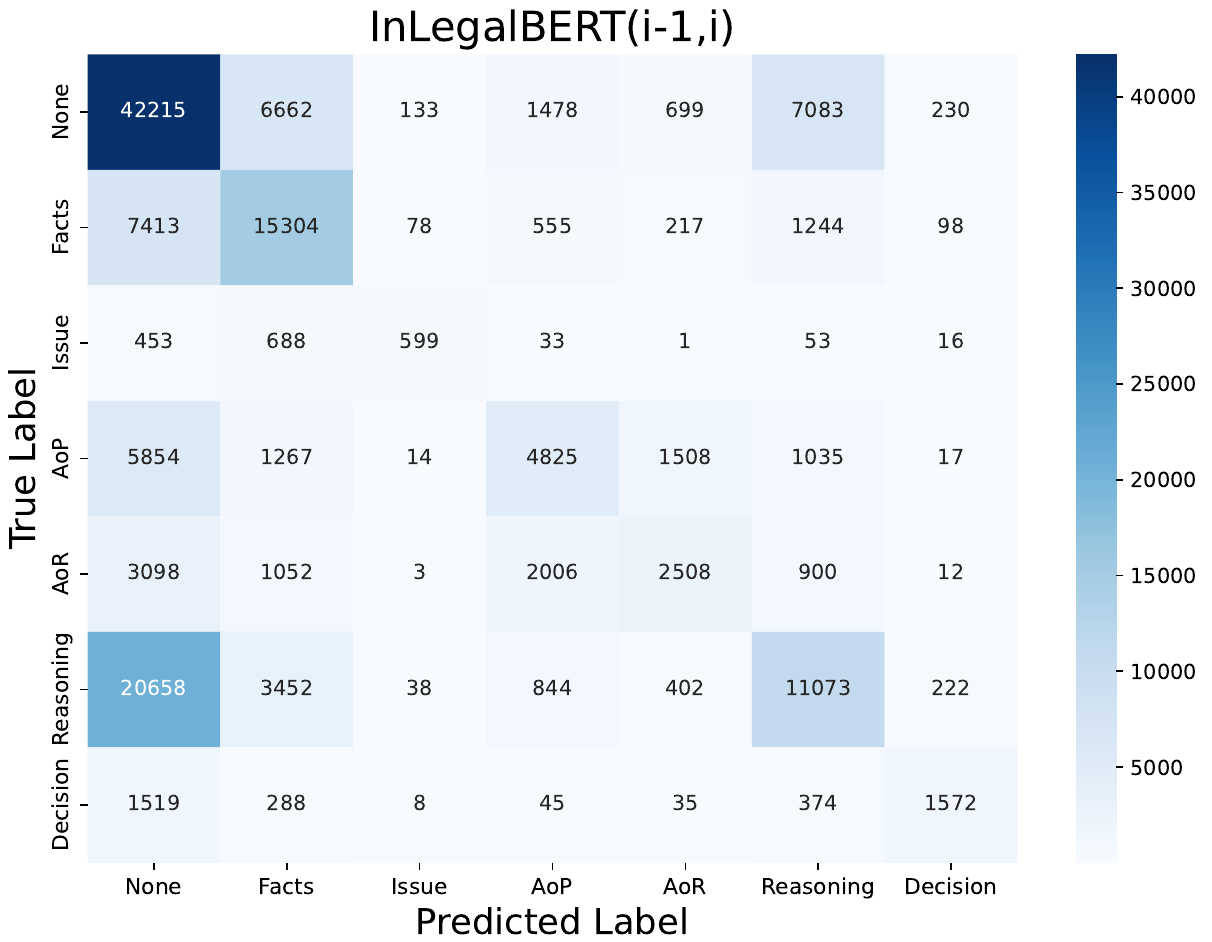}
    \caption{Confusion matrix for rhetorical role classification using InLegalBERT model with the current sentence (i) and the previous sentence (i-1) as input.}
\label{fig:inlegalbert_i-1_i}

\end{figure}

\begin{figure}[ht]
    \centering
    \includegraphics[width=\linewidth]{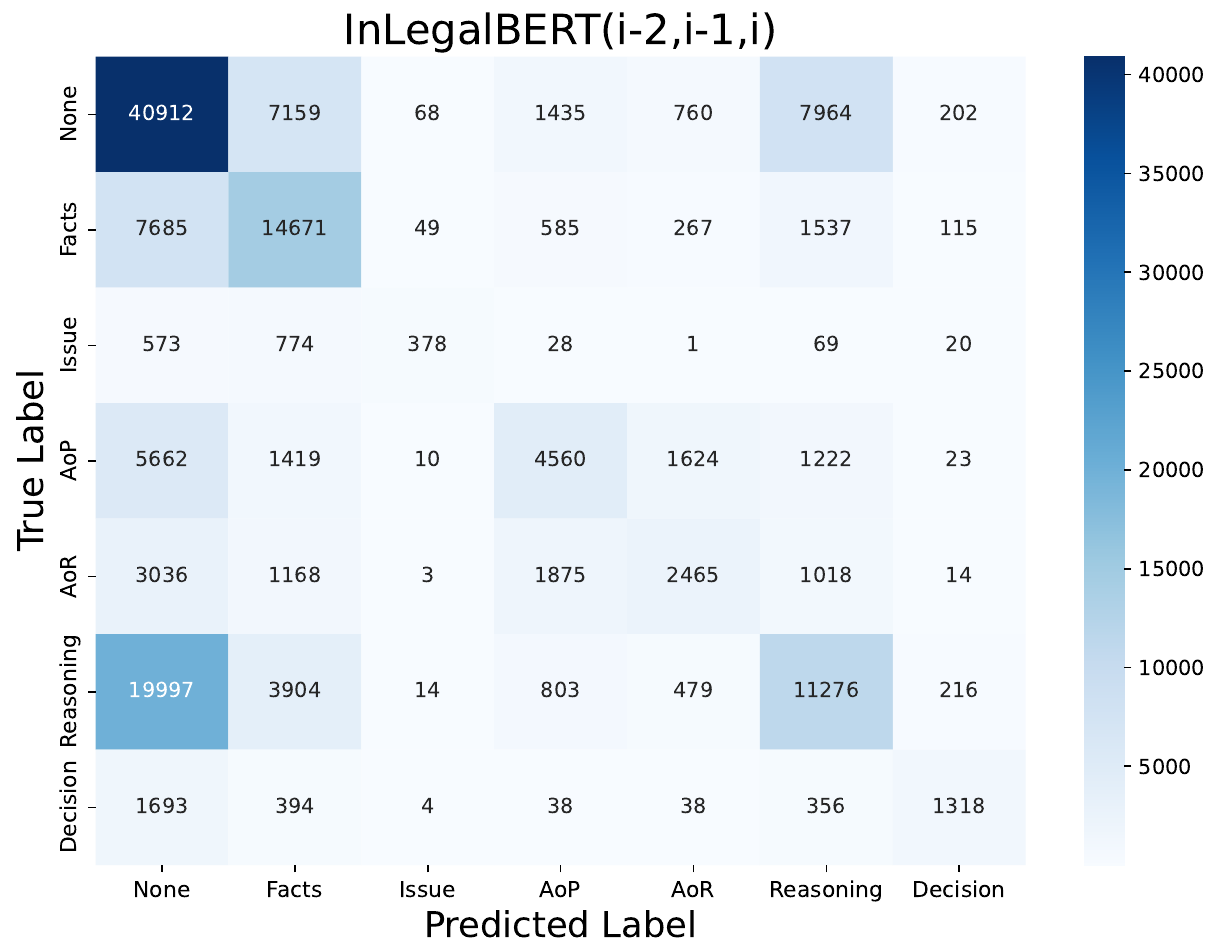}
    \caption{Confusion matrix for rhetorical role classification using InLegalBERT model with the previous-to-previous sentence (i-2), previous sentence (i-1), and the current sentence (i) as input.}
\label{fig:inlegalbert_i-2_i-1_i}

\end{figure}

\begin{figure}[ht]
    \centering
    \includegraphics[width=\linewidth]{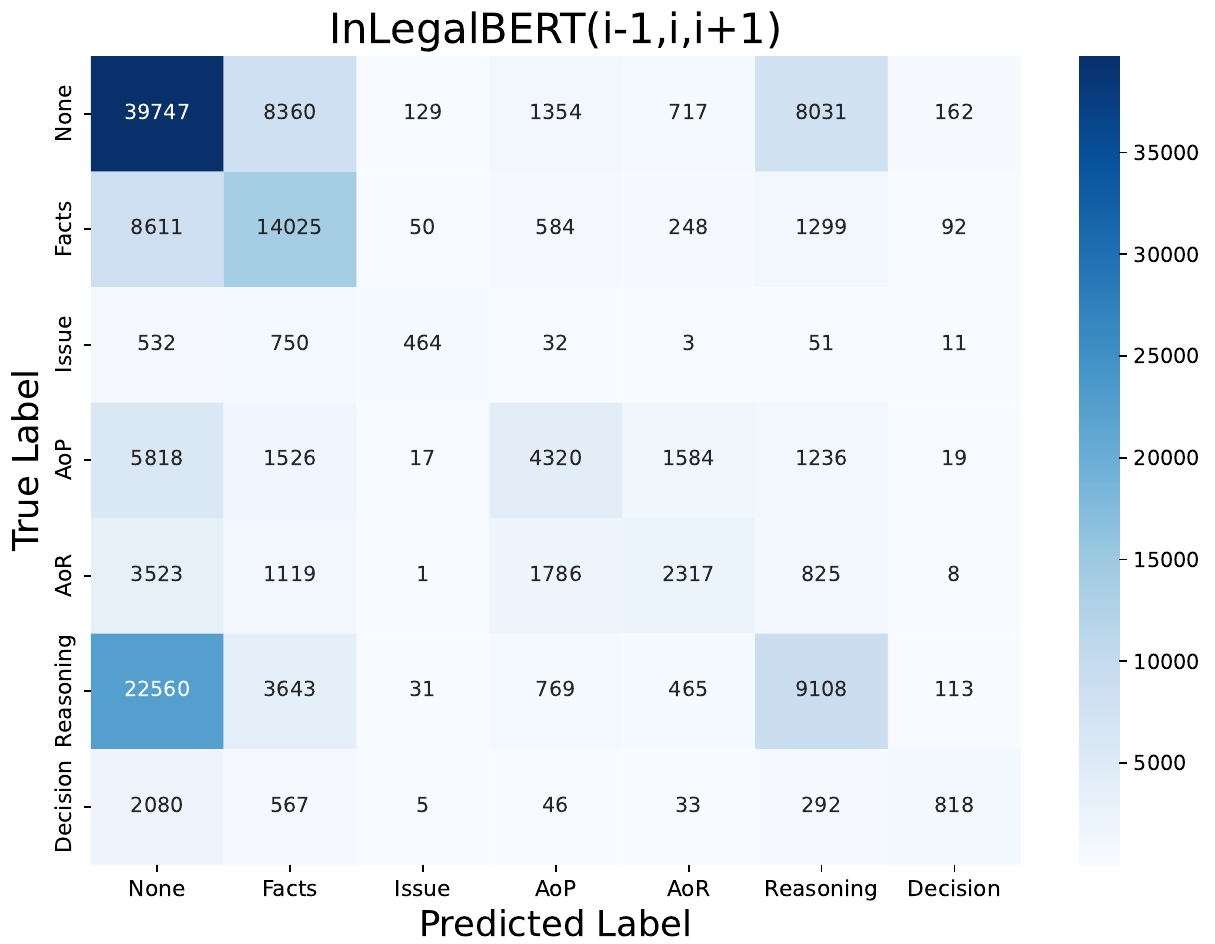}
    \caption{Confusion matrix for rhetorical role classification using InLegalBERT model with the current sentence (i), previous sentence (i-1), and next sentence (i+1) as input.}
\label{fig:inlegalbert_i-1_i_i+1}

\end{figure}

\begin{figure}[ht]
    \centering
    \includegraphics[width=\linewidth]{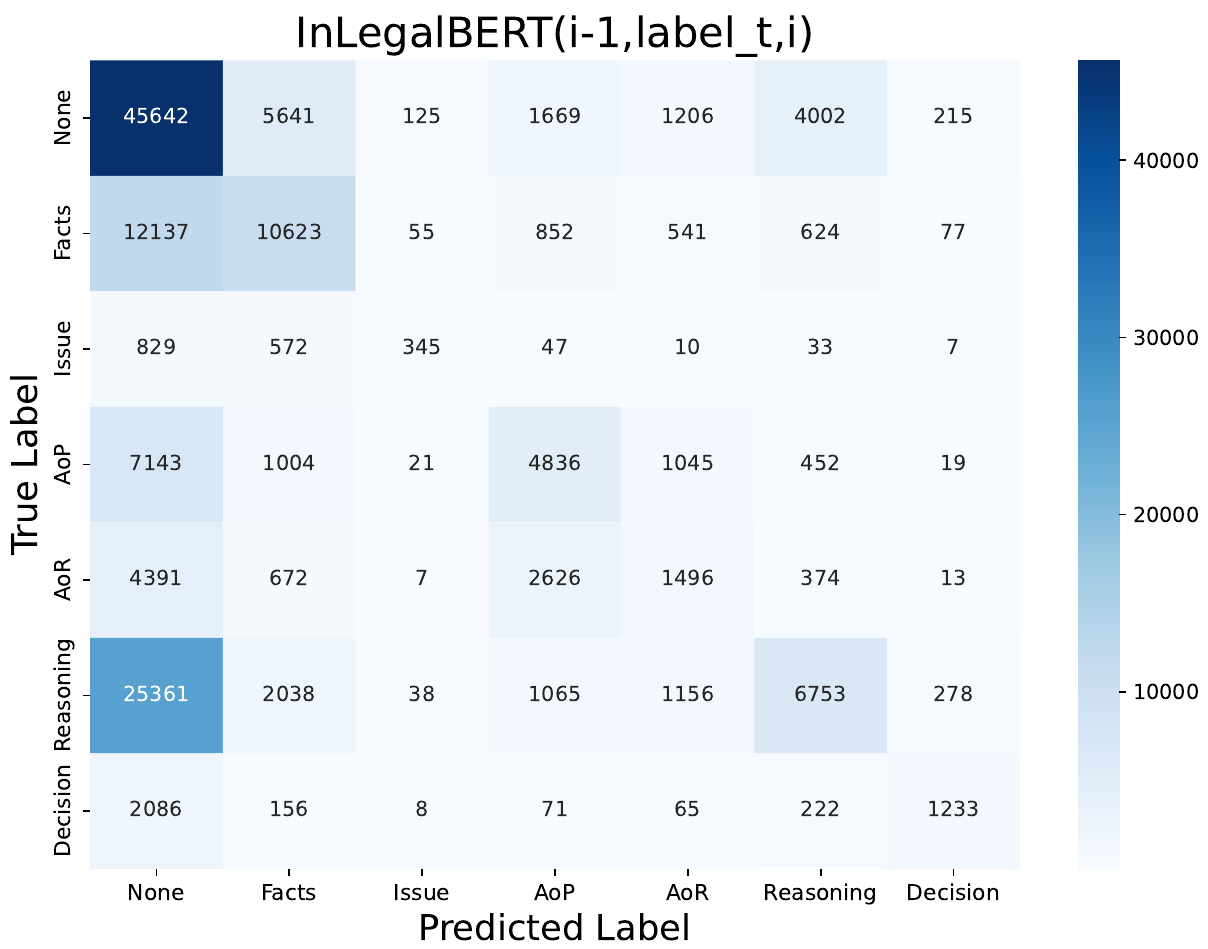}
    \caption{Confusion matrix for rhetorical role classification using InLegalBERT model with the true label of the previous sentence (i-1) and the current sentence (i) as input.}
\label{fig:inlegalbert_label_t}

\end{figure}

\begin{figure}[ht]
    \centering
    \includegraphics[width=\linewidth]{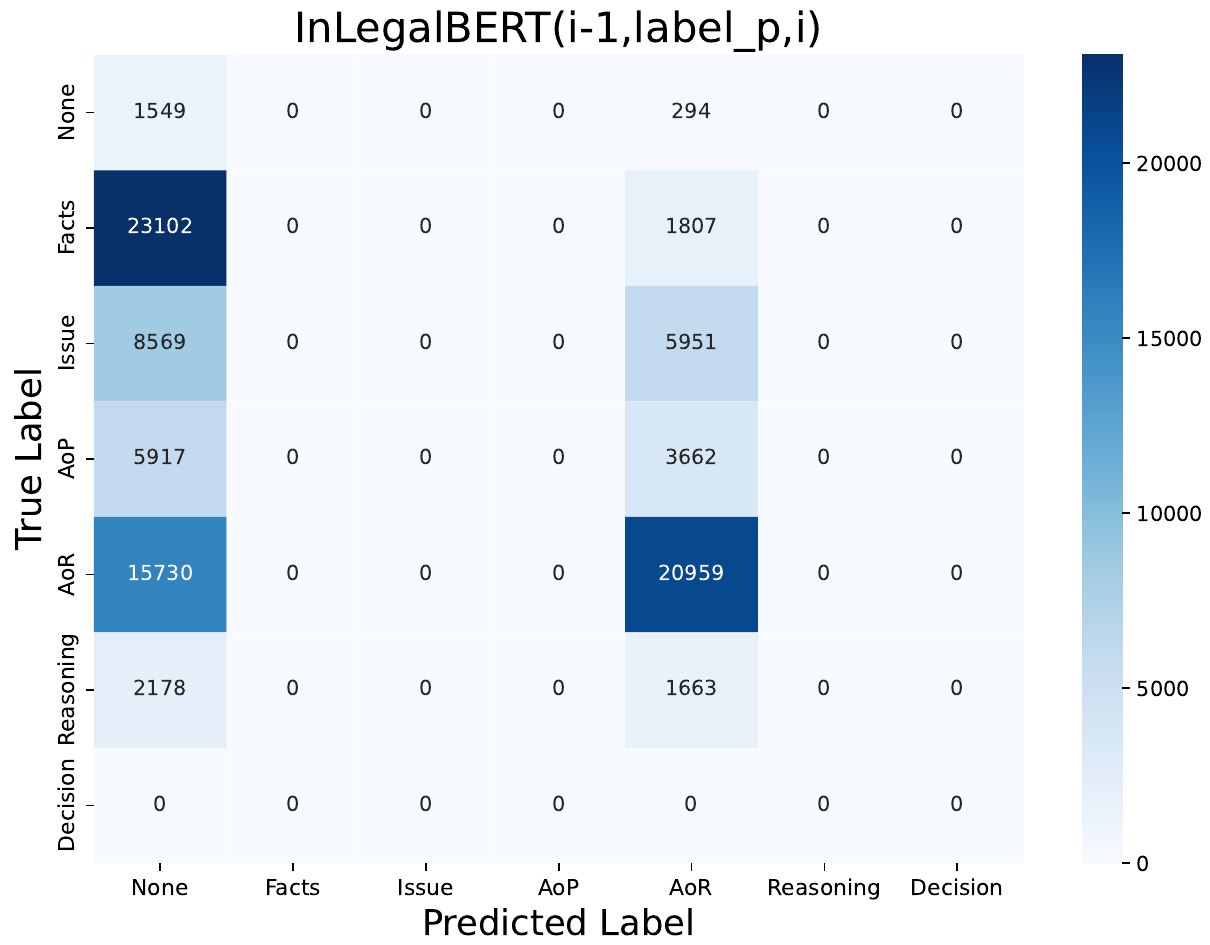}
    \caption{Confusion matrix for rhetorical role classification using InLegalBERT model with predicted label of the previous sentence (i-1) and the current sentence (i) as input.}
\label{fig:inlegalbert_label_p}

\end{figure}

\end{document}